\documentclass{article}

\usepackage[applemac]{inputenc}
\usepackage[T1]{fontenc} 
\usepackage[english]{babel}
\usepackage{amsmath,amssymb}
\usepackage{natbib}
\usepackage{setspace}
\usepackage{authblk}
\usepackage{graphicx,tikz}

\title{ML Interpretability:\\Simple Isn't Easy}

\author{Tim Räz\footnote{University of Bern, Institute of Philosophy,  L\"anggassstrasse 49a, 3012 Bern, Switzerland. E-mail: tim.raez@posteo.de}}

\begin{document}

\maketitle

\begin{abstract}
The interpretability of ML models is important, but it is not clear what it amounts to. So far, most philosophers have discussed the lack of interpretability of black-box models such as neural networks, and methods such as explainable AI that aim to make these models more transparent. The goal of this paper is to clarify the nature of interpretability by focussing on the other end of the ``interpretability spectrum''. The reasons why some models, linear models and decision trees, are highly interpretable will be examined, and also how more general models, MARS and GAM, retain some degree of interpretability. I find that while there is heterogeneity in how we gain interpretability, what interpretability is in particular cases can be explicated in a clear manner.
\end{abstract}

\section{Introduction}
\label{sec:intro}

Machine learning (ML) models, and deep neural networks (DNNs) in particular, are very successful at solving problems both within and outside of science; the latest, spectacular scientific example is the prediction of protein folding \citep{jumpe2021}. However, many of these models are black boxes, and we do not know why they are so successful. As a consequence, the interpretability of ML models -- understanding or gaining insight into how they work -- is an important area of research in computer science. One kind of effort is towards a better grasp of theoretical properties of ML models, and to formulate what is called a theory of deep learning \citep{berne2021,bahri2020}. Another kind of effort is to provide ML practitioners with tools to understand predictions made by the ML models they deploy. This latter effort often runs under the label of explainable AI (xAI, see, e.g., \citealt{adadi2018}). Philosophers have also started to pay more attention to interpretability recently; see \citet{beisb2022} for a survey. Some philosophers have discussed more theoretically-oriented approaches \citep{buckn2019,raez2020a,sterk2021}; there have been proposals for frameworks of explainable AI \citep{zedni2019}; and it has been discussed whether understanding ML models is relevant to their usefulness in application \citep{sulli2019,raez2020}.

The goal of the present paper is to explore the prospects of explicating the concept of interpretability in more precise and unified terms. If we want to increase the interpretability of models such as DNNs, we have to get clear on what interpretability is. Computer scientists like \cite{lipto2016} have noted that interpretability is not a ``monolithic concept''; \citet{doshi2017a} call for a more ``rigorous'' notion of interpretability. Philosophers like \citet{krish2019} concur that interpretability lacks a clear meaning and question whether interpretability is an important problem in its own right.

So far, philosophers have approached interpretability with a focus on black-box models. The present paper takes a different approach. Instead of focusing on the black-box end of the spectrum, where models lack interpretability, the focus here is on the other, interpretable end.\footnote{This approach is inspired by the distinction, stressed by \citet{rudin2019}, between designing ``inherently interpretable'' models as opposed to applying xAI methods to opaque models.} In computer science, interpretability has a tradition that predates the recent ascent of DNNs. The present paper follows the traces of this tradition in order to get a clearer picture of what interpretability is. I will examine four cases of ML models that are interpretable (to a certain degree), which properties make these models interpretable, and how a higher degree of generality affects interpretability in two of the four cases.\footnote{There are many useful discussions of interpretability in \citet{hasti2009}; the four cases discussed below can be found there. See \citet{rudin2019,rudin2021} more background on interpretable models, and \citet{molna2020} for a book-length introduction, including more examples of interpretable models.} The upshot will be that while interpretability is heterogeneous, it is possible to explicate, in a reasonably clear manner, what it means to have a certain degree of interpretability for a certain class of models. 

Section \ref{sec:prelim} introduces important aspects of ML models, and spells out how the debate on (scientific) understanding will be put to work to clarify interpretability. In section \ref{sec:int-reg}, two regression models commonly held to be interpretable, linear models and (binary) decision trees, are discussed. I compare and contrast the properties that make these models interpretable in order to identify core properties of interpretability. Section \ref{sec:general-inter} turns to the question of how interpretability scales with the generality of models. I examine two regression models, MARS and GAMs, that retain a certain degree of interpretability, comparing, again, the properties that make the two models interpretable. Section \ref{sec:discussion} discusses general lessons about interpretability to be gleaned from the four cases. I conclude in section \ref{sec:conclusion}.

\section{Preliminaries}
\label{sec:prelim}

\subsection{Aspects of ML}

The focus of the present paper is on supervised learning \citep{hasti2009}; other paradigms of machine learning like unsupervised learning or reinforcement learning are neglected. In supervised learning, we start with a dataset $\mathcal{D} = \{(x_i, y_i), i = 1...n\}$, sampled from an unknown distribution $P(X, Y)$, with instances $x_i$ of $X$ (inputs), labeled by instances $y_i$ of $Y$ (outputs). The variables $X$ and $Y$ can be continuous or discrete. Here, the focus is on regression problems, i.e., both $X$ and $Y$ are assumed to be continuous. The goal of supervised learning is to find a function $f(X) = \hat{Y}$ that approximates the dataset $\mathcal{D}$ such that predictions $f(x_i) = \hat{y_i}$ are close to $y_i$ according to some loss function. The goal of finding $f$ is achieved by specifying a model $M$ that computes $f$, and an optimization procedure, or learning algorithm, that adapts the parameters of the model $M$ in a learning process, such that the model approximates the relation between the $x_i$ and the $y_i$ in $\mathcal{D}$. In what follows, $f$ will be assumed to be a mathematical function, and variables to be ranging over sets; the probabilistic perspective, in which $X, Y$ are random variables and $f$ a distribution, will be neglected.

The concept of interpretability explored in this paper  -- the degree to which we understand an ML model -- focuses on some aspects of supervised learning, while bracketing others. First, we can try to understand an ML model itself, or we can try to understand something about the world with an ML model. In the case of understanding \emph{with} a model, the model plays an instrumental role in understanding something about the system from which the data $\mathcal{D}$ is sampled. In the case of understanding \emph{of} a model, we are trying to understand the model itself. Here the focus is on understanding \emph{of} a model; the question whether an ML models faithfully captures aspects of the world is bracketed.\footnote{In the philosophical literature, it has been discussed to what extent understanding \emph{of} a model is relevant to understanding \emph{with} a model, which seems the primary concern. \citet{sulli2019} argues that our current understanding of ML models is sufficient to use them to understand the systems modeled; this has been challenged by \citet{raez2020}.} Second, we can try to understand how a model arises through training, or we can try to understand a fixed trained model; see \citet{raez2020a} for an example of the former. Here the focus is on the latter, i.e., understanding a fixed, trained model, or a family of such models. Finally, we can try to understand the inner workings of a model, or we can try to understand the function $f$ computed by the model. Here the focus is on understanding the predictor function $f$. Understanding the inner workings of a model, its algorithmic properties etc. is important, and sometimes, there is no absolute distinction between model and predictor. Here, the inner workings of a model will only be considered insofar as this serves the ultimate purpose of understanding the function $f$.

In short, the kind of interpretability to be investigated here, dubbed \emph{functional interpretability}, is concerned with understanding the predictor function $f$ computed by an ML model. This notion has been discussed in the philosophical literature on ML, in particular in the normative framework for xAI proposed by \citet{zedni2019}. In Zednik's terminology, the notion to be discussed here is understanding \emph{what} the predictor is doing, as opposed to a notion of understanding \emph{how} (in terms of process) or \emph{why} (in terms of a representational relation between model and world). In terms of the kinds of transparency distinguished by \citet{creel2019}, the notion to be discussed here is a variety of functional transparency. Functional interpretability concerns the behavior of a predictor function as a whole, and thus requires a \emph{global} kind of understanding, as opposed to local notions, which focus on understanding (or explaining) single predictions; the latter case is the usual setting of xAI methods.\footnote{Functional interpretability is thus a more general notion that the explanation of particular predictions, as analyzed by, e.g., \citet{watso2021,zeril2022}.}
 
Why is functional interpretability important? First, it is important because the main goal of supervised learning is to make predictions; therefore, it is crucial to understand how a model is behaving for various inputs. While understanding what an ML model predicts will not tell us everything about that model -- it is not sufficient for full-blown understanding -- it is arguably necessary: we do not understand an ML model unless we know what it does. Second, understanding predictions is probably the aspect of ML models that affects most stakeholders -- it is not only of interest to model builders, or model users, but also to decision subjects and policymakers. Therefore, a clear concept of understanding this aspect of ML models is of high practical relevance.

\subsection{Interpretability and Scientific Understanding}
\label{sec:intro_understanding}

In the last section, the aspect of ML models we want to understand were specified, viz. the function $f$. This section discusses what \emph{understanding} means. To do so, relevant literature on understanding from philosophy is reviewed. Understanding is arguably one of the central goals of science \citep{deregt2005}, and has been discussed in various guises in philosophy of science and epistemology. For a long time, understanding was seen as a mere psychological by-product of explanation \citep{woodw2014}; only more more recently has it been recognized as an achievement in its own right; see \citet{baumbergeretal2017} for a survey and \citet{beisb2022} for a discussion of the relation between understanding and intelligibility.

The relation between understanding and interpretability to be used here is summarized in the following slogan: A model with high interpretability is a model for which the degree of understanding is high. Here, understanding is taken to have several distinct but related dimensions (see also \citealt{wilkenfeld2017}). First, we, the agents who want to understand a model or function, need to be able to \emph{grasp} this representation; it needs to be \emph{intelligible} to us. Intelligibility should not be a purely subjective matter, it does not reduce to a sense of understanding \citep{trout2002}. Intelligibility, the grasping of a representation, should be explicated such that grasping can be taught, acquired, and verified an intersubjective manner. There have been different proposals in the literature for how to do this. One proposal is that a representation is intelligible to the extent that it is possible for an agent to reason about the representation, manipulate the representation, and/or use it to make counterfactual inferences \citep{kuorikoski2015}. A second proposal is that a representation is intelligible to the extent that an agent can anticipate qualitative consequences of the representation, without calculations or quantitative inferences \citep{deregt2005}. One particular way of anticipating qualitative consequences of a representation is through visualization \citep{dereg2014}. Below, the cases will be analyzed with the help of these proposals. 

A second important issue is whether understanding is taken to be categorical -- we either have understanding, or we do not -- or \emph{graded}, that is, understanding comes in degrees; see \citet{Baumberger2019,jebei2020}. Here a graded notion of understanding will be used. A further important dimension of understanding is \emph{accuracy}, that is, to what extent a representation allows us grasp something about the system that is represented. Here, however, the question of accuracy can be bracketed, because the focus is on understanding an aspect \emph{of} an ML model itself, not understanding \emph{with} am model, i.e., whether the model provides an adequate representation of something in the world.

In sum, the object we want to understand is the function $f:X \rightarrow Y$ computed by a trained ML model $M$. The function $f$ can be represented in different ways, e.g., by an algorithm that specifies how to compute outputs $Y$ from inputs $X$. However, we are not aiming to understand the computations, we are only interested in the input-output relation. Understanding means grasping this function (to a certain degree), e.g., by observing how manipulating the input changes the output, or by anticipating how the function behaves qualitatively. This proposal, which is quite general and vague at this point, will be refined in view of the case studies, to which we now turn.

\section{Interpretable Models}
\label{sec:int-reg}

In this section, \emph{linear models} and \emph{decision trees} are examined; these two models that are considered to have a high degree of interpretability by ML researchers. I will first identify and discuss properties that contribute to the intelligibility of these models. I will then compare these properties and argue that while these two models share some properties that contribute to their intelligibility, their formal properties and the way in which we grasp them are so different as to yield two different paradigms of interpretability.

\subsection{Linear Models}

\begin{figure}[h] 
   \centering
   \includegraphics[width=60mm]{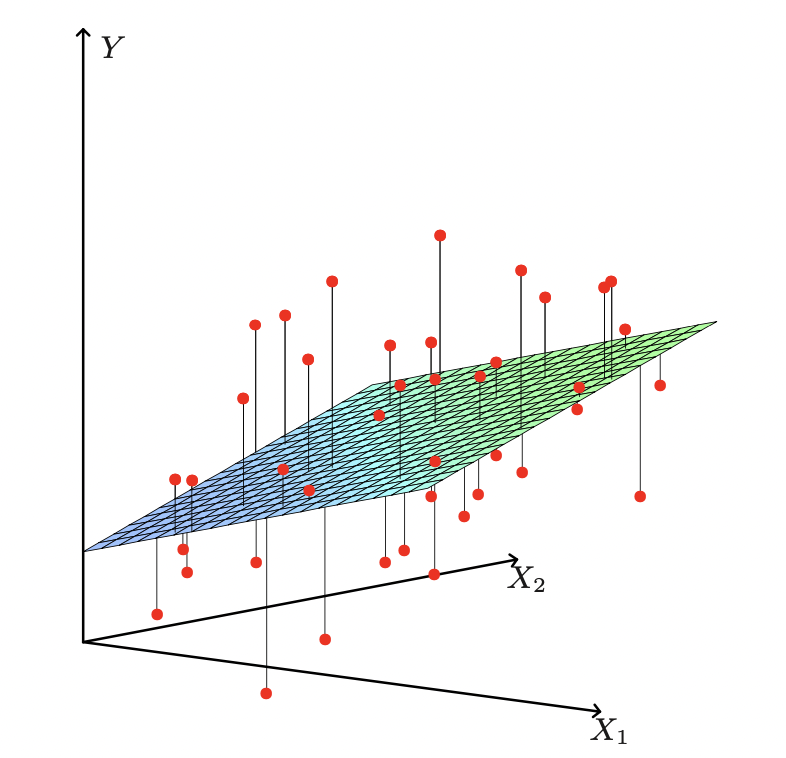} 
   \caption{Linear model with inputs $X_1, X_2$ and output $Y$; the red dots are the data points to be approximated by $Y = f(X_1, X_2)$. From \citet{hasti2009}, © by Hastie, Tibshirani \& Friedman.}
   \label{fig:linear}
\end{figure}

Linear models are an important class of interpretable models. In ``Elements of Statistical Learning'' (ESL), a standard textbook, we find the following (typical) description: ``[Linear models] are simple and often provide an adequate and interpretable description of how the inputs affect the output'' \citep[p. 43]{hasti2009}. A linear model of $n$ (continuous) input variables $X = (X_1, ..., X_n)$ and one continuous output $Y = f(X)$ is of the form

\begin{equation}
f(X) = \beta + w_1X_1 + w_2X_2 + ... + w_nX_n, \label{linear}
\end{equation}

where we have added the intercept $\beta$.\footnote{Strictly speaking, the \emph{function} $f$ is affine rather than linear due to the $\beta$ term, but I will ignore this issue here.} The model parameters $\beta, w_1, ..., w_n$ are to be estimated in the learning process. Figure \ref{fig:linear} provides an illustration.

Why are linear models considered to be interpretable? Let us collect and discuss some relevant properties.

\begin{enumerate}

\item We can assign a simple geometrical meaning to a linear model: it corresponds to a hyperplane over the input space; see figure \ref{fig:linear}. Additionally, such hyperplanes can be easily visualized for one- and two-dimensional inputs. However, models with four or more variables cannot be fully visualized, and it is implausible that linear models of four or more variables are fully non-interpretable.

\item We can assign an intuitive meaning to the model parameters: The $w_i$ correspond to the strength with which the input $X_i$ is weighted in the computation of the output. For linear models, the model parameters directly translate into how the predictor function $f$ behaves.\footnote{\citet{lipto2016} calls this \emph{decomposability}. This property distinguishes linear models from, say, DNNs, where the relation between model parameters and output is not straightforward.}

\item Linear models are additive, which means that there are no interactions between input variables. Once we know the contribution of individual features, the overall output is a sum of these contributions, i.e., additive. This is a weak form of decomposability.\footnote{Additivity is a weaker property than linearity; not all additive models are linear; see the discussion of generalized additive models in section \ref{sec:gams} below.}

\item For linear models, the same change in an input $dX$ leads to the same change in the output $dY$ everywhere. Or, put differently, the local shape of the function is also its global shape. This property uniquely singles out linear models.\footnote{This is true because linear models are the only models with constant derivatives everywhere. Note that we subsume constant functions under linear ones because we do not require the intercept $\beta$ to be zero.} 

\end{enumerate}

This list of different reasons why linear models may be interpretable, while reasonable, may be incomplete, and some of the properties are not mutually exclusive. The first property suggests that linear models are graspable through visualization, while the fourth may be recast in terms of grasping through local manipulation. The second and the third property suggest that linear models are particularly intelligible because of the \emph{form of the predictor function}. It should be noted that there are efficient and well-known procedures to fit a linear model to data. This makes sure we are able to find a linear model in the first place. Also, given any input, we can compute the corresponding output of the model in an efficient manner.\footnote{\citet{lipto2016} calls this \emph{simulatability}.} 

It is not universally accepted that linear models are interpretable \emph{tout court}. \citet{lipto2016} states several reasons why linear models are only interpretable with some qualifications. First, if the input space is high-dimensional, then carrying out computations, be it finding a model or computing particular outputs, becomes harder and harder. What is more, it may also be hard to gain an overall picture of how inputs affect outputs for high-dimensional models, and we may lose our overall grasp of the model's behavior, even if we understand how linear models work in principle.

These qualifications are reasonable. However, they do not apply to linear models exclusively, but are a general property of interpretability: The interpretability of models decreases as the dimension of the input space increases.\footnote{To counteract a loss of interpretability in higher dimensions in the linear case, one can construct sparse models (models with dependence on few variables), e.g. using LASSO regularization, and use variable selection methods; one purpose of these techniques is to make high-dimensional models more interpretable, cf. \citet[Ch. 3]{hasti2009}.} This, of course, is relatively vague, and it may be hard to make this more precise, but it seems natural to say that, generally speaking, the \emph{degree} of interpretability decreases as the dimension of the input space increases. This is compatible with the possibility that we may lose our grasp of a certain kind of model above a certain threshold of the input dimension.

\citet{lipto2016} raises further points: linear models are fragile in that they sensitively depend on the selection of input variables, and also on pre-processing data, and, in order to get a degree of accuracy from linear models that is comparable to more complex models, linear models require pre-processed or pre-engineered features, which, in turn, may compromise interpretability. These arguments are valid in substance; the sensitive dependence on variable selection is a real problem. However, these are issues of understanding \emph{with} the model, or how the model and the world are related, which are bracketed here. Importantly, even in cases where stability is not an issue, we should still have a clear notion of understanding a given predictor function. Spelling this out is worthwhile whether or not model inference is stable.

\subsection{Decision Trees (CART)}

\begin{figure}[h] 
   \centering
   \includegraphics[width=80mm]{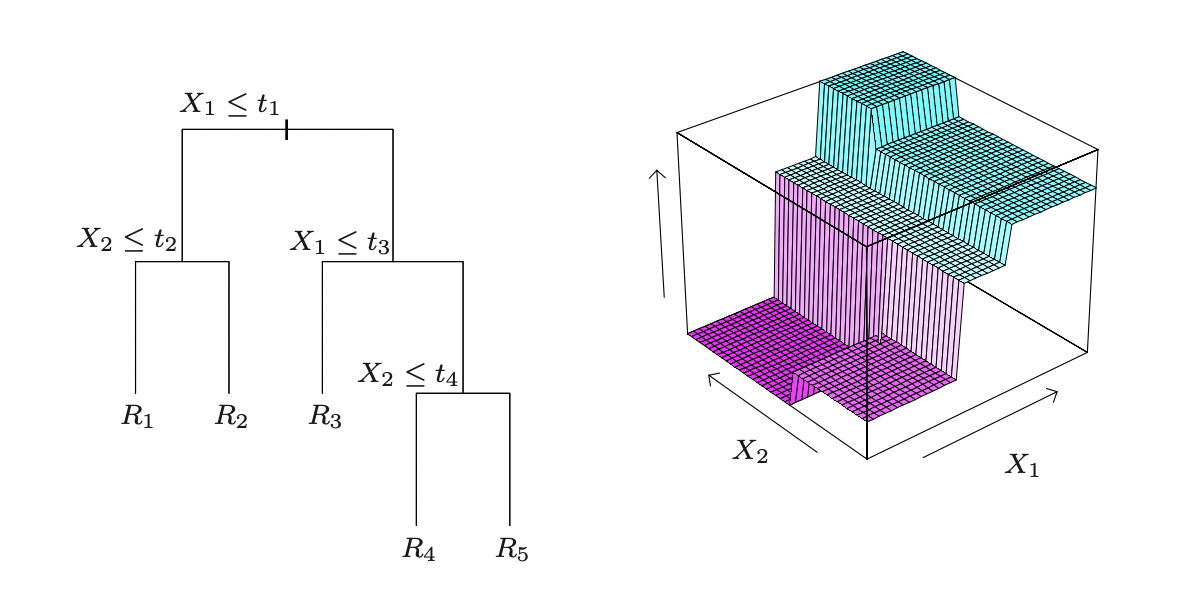} 
   \caption{CART: A binary regression tree in two variables (left) and the corresponding regression function (right).  From \citet{hasti2009}, © by Hastie, Tibshirani \& Friedman.}
   \label{fig:CART}
\end{figure}

The second family of interpretable models are decision trees. Here we look at a simple kind of decision tree, so-called \emph{classification and regression trees} (CART). Decision trees are considered to be interpretable by many computer scientist. \citet{hasti2009} write: ``Tree-based methods partition the feature space into a set of rectangles, and then fit a simple model (like a constant) in each one. They are conceptually simple yet powerful'' (p. 305).\footnote{See also \citet{rudin2021,lipto2016} for similar assessments.} Here is a short description of CART.\footnote{The presentation here follows \citet[Sec. 9.2.]{hasti2009}.} Here is how these trees are constructed. CART are recursive, binary decision trees that correspond to partitions of the input space into regions. The regions are obtained by recursively splitting the range of variables; see figure \ref{fig:CART} for an illustration.

We start with training data $\mathcal{D}$, defined for variables $X = (X_1, ..., X_n)$ and $Y$. In the first step, we search through all possible partitions of the input space into two regions $R_1, R_2$. The two regions have the form $R_1 = X_i \leq t; R_2 = X_i >t$, that is, we split each variable $X_i$ at values $t$ of datapoints in $\mathcal{D}$. We choose the variable $X_i$ and the split $t$ such that, if we calculate the average value of the data in the two regions, and compare the result with the data, we get a minimal empirical loss. The recursive procedure is continued in the different regions until a certain tree size is reached; afterwards, the tree is pruned (internal nodes are collapsed) until a given tradeoff between model complexity (size of the tree) and predictive accuracy is satisfied. The resulting predictor function $f:X \rightarrow Y$ can be written as follows:

\begin{equation}
f(x) = \sum_i c_i \cdot I(x \in R_i). \label{tree_predictor}
\end{equation}

Here, $R_i$, $i = 1 ... m$, are the regions of the partition, $c_i \in Y$ the prediction values in these regions (the average value of $\mathcal{D}$ in region $R_i$), and $I$ the indicator function: $1$ for $x \in R_i$, and $0$ else.

What are the properties that make decision trees (CART) interpretable?

\begin{enumerate}

\item CARTs have a simple geometrical interpretation, they correspond to functions that are constant over (simple) regions of the input space; see figure \ref{fig:CART} above. CARTs also allow for intuitive visualizations as trees, and the corresponding partitions can also be visualized, at least for low dimensions. The last property is very restrictive as a necessary requirement for interpretability, as in the linear case.

\item The regions $R_i$ with constant predictions have a simple description in terms of the input variables. \citet{hasti2009} consider CART to be particularly interpretable for this reason. They note that if we were to consider any (rectangular) partition of the input space, the resulting regions could be complicated to describe. This problem is solved by using recursive binary partitions.\footnote{``A key advantage of the recursive binary tree is its interpretability. The feature space partition is fully described by a single tree. With more than two inputs, partitions [...] are difficult to draw, but the binary tree representation works in the same way'' \citep[p. 305]{hasti2009}.}

\item The prediction of CART is based on a sequence of binary decisions, which is easy to grasp. This means that the model represents a simple prediction process.\footnote{``[The tree representation] is also popular among medical scientists, perhaps because it mimics the way that a doctor thinks. The tree stratifies the population into strata of high and low outcome, on the basis of patient characteristics'' \citep[pp. 305]{hasti2009}.} Note that this is a property of how the model processes the input, whereas the goal here is to characterize the interpretability of the resulting predictor function.

\item Trees automatically select ``salient'' input variables that ``explain'' most of the variance.

\end{enumerate}

According to these properties, the interpretability of decision trees is closely tied to their representation. The key element that needs to be grasped in the case of a decision tree is the \emph{representation of the partition} associated with it. In the case of CART, the partition is grasped through the binary tree, which provides us with simple descriptions of the partition regions $R_i$: Every region is characterized through the sequence of splits at the internal nodes leading to $R_i$. This description is implicit in the form of the predictor function (\ref{tree_predictor}) via the regions $R_i$. For decision trees, visualization is also important, but it plays a different role than in linear models: the visual representation of the tree helps us grasp the structure of the partition associated with it. Also, a tree visualization (cf. left of figure \ref{fig:CART}) may help us grasp a function of more than two variables, whereas visualizations of predictor functions (cf. right of figure \ref{fig:CART}) are not available.

Decision trees, like linear models, are not interpretable without qualification. First, Lipton's argument about the dimension of the input space applies to decision trees as well. An important difference between linear models and decision trees is that we may be able to grasp a small decision tree even if the input space is large. A small decision tree is a tree with few splits, which means that the variables that are not split do not contribute to the prediction and can be ignored. Note that the same can be said about sparse linear models.

Second, decision trees can overfit the data if they become too large. If no stopping condition is used, and the tree is allowed to grow indefinitely, a tree can fit any (finite) dataset perfectly. This means that if the size of trees is not limited, they are not intrinsically interpretable. This feature of decision trees is not shared by linear models, which do not overfit the data to the same extent. Thus, the interpretability of decision trees has to be qualified: It applies to small trees only. This is, again, a point where a graded notion of understanding is important: With a categorical notion of interpretability, it would be necessary to say what ``small'' means, whereas on a graded notion, we can say that interpretability decreases as the tree grows. Third, decision trees can be sensitive with respect to small changes in the training data; see, e.g. \citet[p. 312]{hasti2009}. As in the case of linear models, this problem is outside the scope of the concept of functional interpretability explored here.

\subsection{Linear Models and Trees: Two Paradigms of Interpretability}

Now we put the analysis of linear models and decision trees to work and examine whether we can find a common explication of the interpretability of these two kinds of models. If interpretability is a monolithic concept, we should be able to identify common properties of these two highly interpretable models, and spell out what makes a model highly interpretable on this basis. If interpretability is not monolithic, but heterogeneous, we may still be able to characterize the main features that make these kinds of models interpretable separately, but without much overlap between the main properties.

One way in which we might explicate interpretability is via common mathematical properties of predictor functions. This, however, does not seem to work: If we want to assign a high degree of interpretability to both linear models and decision trees, then highly interpretable predictor functions are not necessarily a) linear, b) differentiable or smooth, c) continuous, d) monotone, because small decision trees lack all of these properties in general.\footnote{Here I assume that the domain of trees is the feature space before the recursive partitioning step.} Thus, there is no straightforward characterization of a high degree of interpretability through mathematical properties of predictor functions. There is a very limited class of functions that belongs to both linear models and decision trees, the functions at the ``intersection'' of the two families: a decision tree with no splits, which corresponds to a linear model where all coefficients except for the bias term are $0$. These globally constant functions are maximally interpretable. They compress the data into one number, which can be chosen to be the mean value or the median of the data.

Now let us examine the lists properties that make the two kinds of models interpretable. Both linear models and decision trees allow for visualization in low dimensions, and it makes sense to say that both models have a high degree of interpretability from an intuitive point of view. However, many other aspects of the interpretability two kinds of models are different. First, linear models are primarily grasped through the \emph{form of the predictor function}: the form of the linear function shows that the (constant) contributions of individual variables to the output can be considered separately and contribute additively to the output. The predictor function of a decision trees, on the other hand, is grasped through the \emph{form of the partition}, which is mirrored in the structure of the tree. If you want to grasp the predictor function of a tree, you need to grasp, first, how the tree partitions the entire input space (through the visualization of the tree), and, second, you need the values assigned to the regions of the partition to get outputs. A second, related dissimilarity is that linear models allow for local-to-global inference: once you know how a linear model behaves at one point, you know how it behaves everywhere. You can grasp a linear model through local manipulation. This is not true for decision trees. The behavior of the predictor function of a tree in one region does not tell you anything about its behavior elsewhere. Another way of putting it is that linear function can be grasped bottom-up, while a decision tree has to be grasped top-down -- you start with the partition and proceed to the values in the regions. A third dissimilarity is that decision trees are not intrinsically interpretable. To make them interpretable, a tradeoff between accuracy and simplicity has to be chosen. Linear models, on the other hand, intrinsically do not overfit the data.

These dissimilarities suggest that linear models and decision trees belong to two different paradigms of interpretability. On the linear paradigm, interpretability primarily hinges on the form of the predictor function, and a high degree of interpretability results from the fact that the form of the function -- constant weights and additive form -- is easy to grasp. On the tree paradigm, interpretability hinges on the way in which we partition the input space, and a high degree of interpretability results from the fact that the form of the partition is simple (description/visualization as a binary tree). Thus, even on the minimal kind of functional interpretability considered here, interpretability is not a monolithic concept. The point is that we grasp the predictor functions in two different ways: via the representation of the function itself, and via the representation of the partition found by trees.

How are these findings related to the discussion of grasping in section \ref{sec:intro_understanding}? On the one hand, several of the criteria for grasping from the literature prove to be useful in the present context. We can grasp linear functions through local manipulation, and we can reason qualitatively on the basis of visualizations about both linear models and trees. On the other hand, some aspects of grasping a predictor function identified here do not feature prominently in the literature. Importantly, we grasp both linear functions and trees through the \emph{form} in which they are represented, be it the predictor function or the partition associated with the predictor. The form of the representation plays a key role in our ability to grasp the predictor functions.

\section{Generalized Interpretable Models}
\label{sec:general-inter}

In this section, I turn to MARS and GAMs, two more general models that retain a certain degree of interpretability. This will help understand how the degree of interpretability changes with generality. The comparison between the models will show that a unified explication of interpretability is hard to come by, but that there is a reasonably clear way in which these models are interpretable if considered in isolation. I will also explore the possibility of unifying the linear paradigm and the tree paradigm, and argue that MARS provides a weak form of a unification of these two paradigms, with an explicit tradeoff between them.

\subsection{Multivariate Adaptive Regression Splines (MARS)}

\begin{figure}[h] 
   \centering
   \includegraphics[width=80mm]{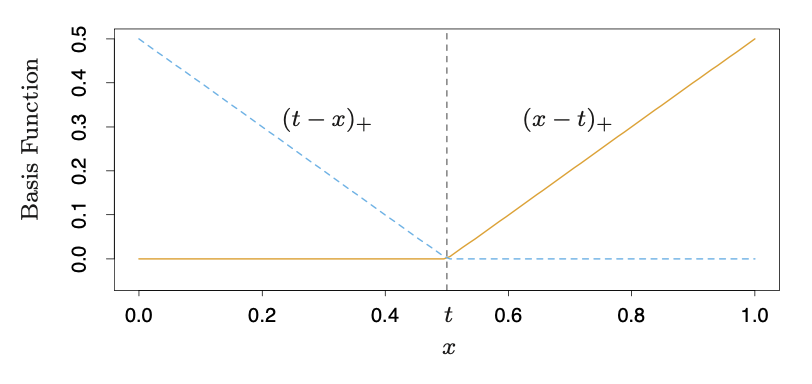} 
   \caption{Pair of basis functions (ReLUs) for MARS. From \citet{hasti2009}, © by Hastie, Tibshirani \& Friedman.}
   \label{fig:ReLU}
\end{figure}

``Multivariate Adaptive Regression Splines'' (MARS), a kind of regression model, combines (stepwise) linear regression and the CART regression tree model.\footnote{The presentation here follows \citet[Sec. 9.4.]{hasti2009}.} The model is built in two stages, a building stage and a pruning stage, similar to regression trees. In the first stage, the model is built recursively from a set $\mathcal{C} = \{(X_i -t)_+, (t - X_i)_+\}$ of pairs of piecewise linear functions in one variable, with knots $t$ at the data points, see figure \ref{fig:ReLU}.

\begin{figure}[h] 
   \centering
   \includegraphics[width=80mm]{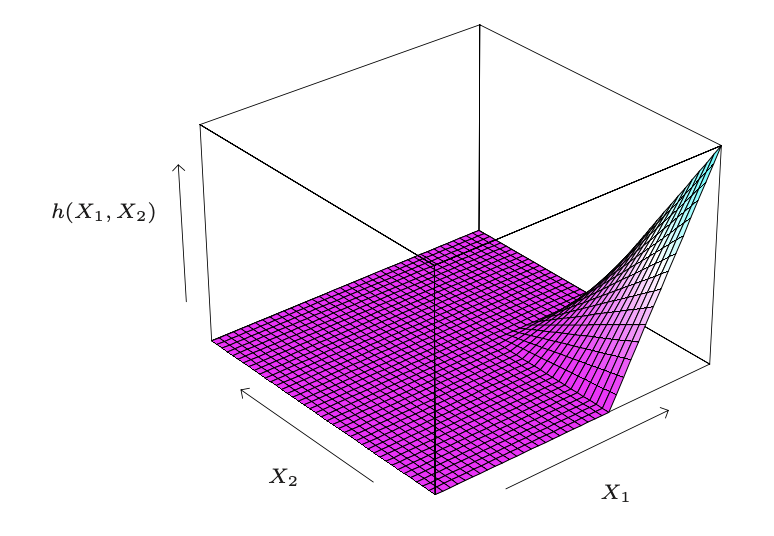} 
   \caption{A prediction function resulting from the MARS procedure; the function is $h(X_1, X_2) = (X_1-x_{51})_+\cdot(x_{72}-X_2)_+$; $x_{51}$ and $x_{72}$ are data points. From \citet{hasti2009}, © by Hastie, Tibshirani \& Friedman.}
   \label{fig:MARS}
\end{figure}

The starting point is the constant function. At each step of the building stage, two new terms are added to the model.\footnote{MARS has several meta-parameters at the building stage that determine what kind of predictor function is constructed. First, an input variable can only feature once in a term. This means that there are no non-linear terms of one variable. Second, the degree of interaction terms, that is, the size of products can be chosen. If the upper limit is $2$, then interactions are quadratic; if the limit is $1$, then the model is additive.}  All possible products of pairs from $\mathcal{C}$ with terms already in the model are considered, and the two terms that lead to the largest decrease of the empirical error are added to the model. At the end of the building stage, the model consists of a sum of products of piecewise linear functions:

\begin{equation}
f(X) = \beta_0 + \sum_{m=1}^M\beta_mh_m(X), \label{MARS_pred}
\end{equation}

where each of the functions $h_m$ is one of the functions in $\mathcal{C}$, or a product of such functions. The building stage terminates once the sum contains a preset number of terms. After the first stage, the model will usually overfit the data. In the second, pruning stage, the model is reduced by sequentially eliminating the term whose removal increases the empirical error the least. This yields a sequence of models of decreasing sizes. From this sequence of models, the final model is chosen through what is called generalized cross-validation, which selects the model that presents the best tradeoff between fit and model simplicity. I will discuss generalized cross validation below. An example of a prediction function resulting from the MARS procedure is shown in figure \ref{fig:MARS}.

After this brief account of how MARS works, let us turn to its interpretability, which is one of the main objectives of MARS.\footnote{Other objectives are accuracy, smoothness, and computability; see the introduction of \citet{fried1991}. Note that \citet[Table 10.1., p. 351]{hasti2009} classifies MARS as an interpretable method.} The property of MARS that contributes most to its interpretability is that its predictor function can be represented in a certain way.\footnote{See the discussion in \citet[Sec. 3.5.]{fried1991}, on which the following discussion is based.} The predictor function resulting from the construction process described above is of the form (\ref{MARS_pred}), which is not particularly telling. However, the terms in (\ref{MARS_pred}) can be rearranged to yield what \citet{fried1991} calls an ANOVA decomposition:

\begin{equation}
f(X) = \beta_0 + \sum_{K_m = 1}f_i(X_i) + \sum_{K_m = 2}f_{ij}(X_i, X_j) + \sum_{K_m=3}f_{ijk}(X_i, X_j, X_k) + ... \label{ANOVA}
\end{equation}

$\beta_0$ is the intercept term. The index $K_m = 1$ runs over all functions $h_m$ which contain one function from $\mathcal{C}$; $K_m =2$ runs over all functions that are product of two functions from $\mathcal{C}$, and so on. The functions $f_i(X_i)$ in the first sum consist of the piecewise linear functions from $\mathcal{C}$ in the variable $X_i$ that enter into the model; the functions $f_{ij}(X_j, X_j)$ consist of all products of piecewise linear functions from $\mathcal{C}$ in the variables $X_i, X_j$ that enter into the model, and so on. 

How does the ANOVA decomposition (\ref{ANOVA}) make MARS interpretable? First, the decomposition (\ref{ANOVA}) reveals which variables enter additively into the model (and which ones do not), which pairs of variables enter quadratically into the model (and which ones do not), and so on. Second, at least the additive and the quadratic parts of (\ref{ANOVA}) can be further investigated through visualization. If the degree of interaction is limited to two, this enhances the interpretability of the model significantly. If interaction terms are excluded, then MARS is additive, and thus even more interpretable, because plotting and interpreting one-dimensional functions is easier than interpreting two-dimensional ones.

Let me turn to the question as to how MARS provides a ``unification'' of linear models and regression trees. The stepwise introduction of functions that are zero on part of the domain corresponds to a splitting or ``breaking up'' of the input space. At the same time, MARS generalizes stepwise linear regression, thus using aspects of the linear paradigm. In stepwise linear regression, the number of input variables in a linear model is controlled by only adding those input variables that contribute most to the accuracy of the model. MARS constructs the predictor function one variable at a time in the building stage.\footnote{Building sparse linear models is partially motivated by interpretability; see the discussion of stepwise linear regression in \citet[Sec. 3.3.2.]{hasti2009}.} These are aspects that MARS inherits from the two paradigms in the optimization phase.

The two paradigms appear most explicitly in the model selection during the second phase of construction. In this phase, the best model from the sequence of models $f_\lambda$ is chosen, where $\lambda$ corresponds to the number of terms in (\ref{MARS_pred}). For the choice of the best model, the criterion for generalized cross-validation (GCV) is used. GCV constitutes a tradeoff between predictive accuracy and simplicity. Both the use of many variables and the use of many splits make a model less simple according to the measure of complexity embedded in GCV. This means that the simplicity in MARS is itself not simple, but complex -- it is literally a sum of two different ``parameters of simplicity'', one associated with the linear paradigm, one with trees. In this sense, MARS is an entire family of interpretable models, given by a parameter controlling for a tradeoff between two paradigms of simplicity. What is more, there is no theoretical justification for how to choose this tradeoff. The tradeoff is a pragmatic choice, which can be based on empirical accuracy. MARS thus inherits the two paradigms of interpretability, between which we have to choose.\footnote{The tradeoff in ML between accuracy and simplicity is well-known both in computer science and philosophy; see, e.g., the discussion in \citet{forst1994} of this tradeoff with respect to model selection. The above discussion of GCV does not concern the tradeoff between accuracy and simplicity. GCV constitutes a three-way balance between accuracy and two ways of conceptualizing simplicity.}

\subsection{Generalized Additive Models (GAMs)}
\label{sec:gams}

In this section, I turn to Generalized Additive Models (GAMs)\footnote{The presentation here follows \citet[Sec. 9.1.]{hasti2009}}, a generalization of linear models. While linear models have the form (\ref{linear}), and thus assign constant weights to individual inputs, additive models are more general in fitting smooth functions to individual inputs, while retaining additivity, that is, there are no interactions between inputs. An additive model has the following form:\footnote{Here I gloss over some aspects of GAMs, such as the fact that GAMs may allow the response $Y = f(X)$ to be a smooth transformation of the sum on the RHS of (\ref{GAM}), via a so-called link function.}

\begin{equation}
f(X) = \alpha + f_1(X_1) + ... + f_n(X_n). \label{GAM}
\end{equation}

The $f_i$ may be non-linear, smooth functions, but it is also possible to choose some of the $f_i$ to be linear and some to be non-linear, depending on what is known about the variables; GAMs are modular in this respect.

GAMs can be fit to data using the so-called backfitting algorithm. The idea of backfitting is to choose some initial values for the $f_i$, and then iteratively fit the $f_i$ to the data using a procedure $S(f_i)$ until the fit stabilizes.\footnote{There are different possible choices for the procedure $S$, e.g. so-called cubic smoothing splines. What is important is that there are efficient procedures to find a good and smooth approximation of a non-linear function $f_i$ in one variable.} To elaborate, we first fit $f_1$ to the data in a smooth way, while keeping the $f_j, j \neq 1$ constant, then we fit $f_2$, using the new estimate of $f_1$ and the initial estimates for the remaining $f_i$, and so on; after one round, we start again with $f_1$, using the estimates for the $f_j, j \neq 1$ from the previous round. We continue estimating in this round-robin fashion until the differences between the $f_i$ in two consecutive rounds are below a certain threshold.

Why are GAMs considered to be interpretable? \citet[Sec. 4.3.]{hasti1990}, who invented GAMs in the 1980s, note that GAMs are useful to analyze data. The form of the predictor (\ref{GAM}) yields a generalization of the interpretability of linear models, which is based on additivity: If you change only one input variable, while holding all others fixed, the corresponding change in the output does not depend on the values of the other input variables. Hastie and Tibshirani write: ``In practice this means that once the additive model is fitted to the data, we can plot the [functions $f_i$] separately to examine the roles of the predictors in modelling the response'' (Ibid. p. 88). Thus, the interpretability of GAMs is based on the fact that the predictor has the additive form (\ref{GAM}) by design, and that, as a consequence, it is possible to examine the separate contributions of the $f_i$ to the prediction by examining the plots (visualization) of the $f_i$.\footnote{In the above description, GAMs are a fully automatic procedure. One problem with this approach is that the backfitting algorithm fits all variables, even those that do not contribute much to the output. It is possible to select and exclude variables by hand. GAMs are usually applied in this more interactive way in data analysis. However, there are also automatic procedures for finding sparse additive models \citep[Sec. 9.1.3.]{hasti2009}.}

\subsection{MARS and GAMs: Compare and Contrast}

Which aspects of interpretability do MARS and GAMs have in common, and which are dissimilar? The two models are similar in the way in which they achieve interpretability. In both cases, interpretability is a two stage process. In the first stage, we use that the predictor function takes a particular form, a decomposition that shows what individual variables and low-degree interactions contribute to the overall output. In the second stage, the parts of this decomposition, the summands, are investigated separately, either through visualization (by plotting the individual functions of the decomposition), or by qualitatively investigating higher-degree interactions (in the case of MARS). By looking at the plots of individual functions, we can grasp qualitative aspects of the contributions of individual variables; of course, it is also possible to analyze the component functions in a more quantitative way.\footnote{Note that visualization can be generalized to a certain extent and applied to functions with higher-degree interactions, e.g. through so-called \emph{partial dependence plots}, cf. \citep[pp. 369]{hasti2009}, and also \citet{molna2020}.}

However, the two models are dissimilar in several respects. First, they constitute generalizations in different directions. GAMs are generalizations of linear functions in that they preserve additivity. GAMs place no \emph{a priori} restrictions on the form of the component functions, which can have almost arbitrary form (usually, they are nonparametric smooth functions). MARS, on the other hand, are geared towards discovering (low degree) interactions through the products of piecewise linear functions; this is a generalization in the spirit of trees. However, the component functions in the decomposition of MARS (basically low degree polynomials) are usually much simpler than the component functions in GAMs (nonparametric smooth functions). This facilitates an interpretation of the components of the ANOVA decomposition of MARS through the form of the component functions.

Second, MARS and GAMs differ in the way the predictor function is constructed, and, as a consequence, what the representation of the predictor function tells us. GAMs place the global constraint of additivity on the form of the predictor function. There is no variable selection (at least in the automatic version), and all component functions are treated equally. MARS, on the other hand, is an intrinsically adaptive procedure that automatically selects variables and interactions that are important for prediction. The ANOVA decomposition only contains variables and interactions that are important. The selective nature of the decomposition provides us with more information about the data -- this is absent in GAMs. However, the ANOVA decomposition of MARS depends on the choice of the tradeoff in GCV, and this choice is pragmatic. The variable selection in MARS is at least partially determined by our preference for a more tree-like or a more smooth (linear) predictor function.

\section{Discussion}
\label{sec:discussion}

\paragraph{Patterns of Interpretability}

The above models exhibit some common patterns of interpretability that are worth pointing out. The most important pattern of how we grasp a predictor seems to be a combination of two different means of grasping: First, the predictor function is grasped in virtue of the form of their representation (either the form of the predictor function itself or the form of the domain partition). Second, certain aspects, or parts, of this formal representation are grasped through visualization, or other modes of qualitative reasoning. These two steps are used in sequence, not in parallel; this can be seen most clearly in the case of GAMs and MARS, but also in the case of trees. Thus, one important lesson for interpretability gained here is that we grasp predictor functions not only by looking at their visualizations, but also by looking at their \emph{form}, their symbolic representation.\footnote{There have been few discussions of the importance of the right kind of representation / notation in the context of mathematical explanation, cf. \citet[Ch. 8]{colyv2012}, \citet{raez2018}.}

The importance of visualizations for interpretability is well known; they are an important part of explainability methods in xAI, cf. \citet{molna2020}. However, the fact that the formal representation of predictors (and the domain) is a key ingredient for interpretability is relatively unexplored. This negligence may be related to the fact that little attention has been paid to formal dimensions along which interpretability varies, such as the size of the input space, the degree and complexity of interactions, and the nature of nonlinearities. The examination of the above models helps us to appreciate how these different dimensions interact and how we may have to trade them off against each other.

\paragraph{Four Dimensions of Interpretability}

From the above discussion, we can extract at least four dimensions that contribute to the degree of interpretability of ML models. First, the degree of interpretability decreases as the size of the input space grows. Second, and relatedly, the degree of interpretability increases as the size of the model decreases, e.g. by making a linear model sparse, or by controlling the size of a decision tree. Third, the degree to which we allow non-linearities in one variable affects interpretability. For example, using nonparametric smoothers in GAMs results in less interpretable one-variable functions than piecewise-linear functions in the additive part in MARS. Fourth, interpretability decreases as we allow interactions of higher degree and complexity .

Note that these four dimensions need not provide a complete or unique characterization of \emph{the} degree of interpretability. The relations between the four dimensions of interpretability are complex, and this complexity make it hard to assign a consistent, singular degree of interpretability. To give an example why it is not easy to construct a single degree of interpretability as a function of these dimensions, it may be tempting to assign a degree of interpretability solely based on the degree of interaction terms, arguing that once we move beyond two-degree interactions, we lose the possibility to visualize, and thus the possibility to achieve interpretability (think of MARS and GAMs). This suggests that we should define interpretability in terms of degrees of interactions. This, however, neglects the fact that even relatively small trees can exhibit higher degrees of interaction, while arguably still being interpretable. Trees are interpretable because their interactions are of a very simple kind -- a sequence of binary decisions in the case of CART. Thus, it is a mistake to tie interpretability to a low degree of interaction in a categorical way.

\paragraph{The Benign Heterogeneity of Interpretability}

The comparison of the four models suggests that we should not expect a monolithic notion of interpretability. Even the very limited notion of \emph{functional interpretability} investigated here is heterogeneous, as we saw most clearly in the comparison between the linear paradigm and the tree paradigm. I argued that these two paradigms not only correspond to two kinds of predictor functions, but that they also differ in the way we grasp them. I examined one possibility of unifying the two paradigms in the form of MARS, and argued that MARS does retain aspects of both linear models and trees. However, the two paradigms do not vanish through generalization, but remain encoded in an explicit tradeoff in the GCM criterion. Note that the two paradigms are just one manifestation of heterogeneity. There are similar, subtler differences between the interpretability of MARS and GAMs.

This heterogeneity does not mean that we should not strive for interpretability as a goal in and of itself, \emph{pace} \citet{krish2019}. It also does not mean that interpretability is ``ill-defined'', \emph{pace} \citet{lipto2016}, or that anything goes. Rather, while there are different means to obtain interpretability, once these means are identified in particular cases, we may have a clear sense of what we do understand about the corresponding models. To make an analogy, proving different mathematical proposition is heterogeneous as well, in that different methods may be necessary to establish different propositions. It is comparatively easy to check whether and why a given proof of a proposition is acceptable, and the criteria for checking this are also clear. Similarly, the ways in which we achieve interpretability are heterogeneous, they have to be adapted to particular kinds of models, and they are not yet available for many models. Still, the tools we do have to interpret GAMs, or MARS, are clear and understandable. It is just that they may not be equally useful in the interpretation of other classes of functions or other kinds of ML models.

\paragraph{The Prospects of a Conceptual Analysis of ``Interpretability''}

The idea that interpretability is a graded notion that varies along several dimensions is more general than the more traditional approach of identifying necessary and sufficient conditions for the concept of interpretability. If one is unwilling to give up on the project of identifying necessary and sufficient conditions, it is, in principle, possible to obtain a conceptual analysis on the basis of a graded notion of interpretability by introducing thresholds, i.e., by declaring that a model is interpretable if and only if it is interpretable at least to a fixed degree $d$. This presupposes that such a degree of interpretability is available, and that a principled choice of threshold can be made. The challenge of coming up with a consistent, graded notion of interpretability, noted above, suggests that it will be even harder to find necessary and sufficient conditions for interpretability than to assign consistent degrees of interpretability.

\paragraph{Ramifications for Black-Box Models and xAI}

What are the ramifications of the above discussion for the interpretability of black-box models? If we try to locate black-box models such as DNNs along the four dimensions of interpretability, we have to assign them a low degree of interpretability, because they lie at the low-interpretability end of all four dimensions: They are usually applied in high-dimensional settings (e.g. image recognition), they have a lot of free parameters, and they are not only highly non-linear, but capture high-degree interactions. It is one of the main open puzzles about these models that they are predictively successful despite these properties. One of the main messages of the present paper is that even if we focus on functional interpretability, there is not just one method of grasping a predictor function. Rather, we may need to employ different methods in combination for black-box models, as in the case of MARS and GAMs, where both formal properties of the representation of the predictor as well as visualizations play key roles in grasping. 

Functional interpretability aims at global understanding (of an entire function) and is therefore different from local xAI methods \citep{adadi2018}, which rely on local and linear approximations in combination with visualizations. We can nevertheless glean some lessons from the above discussion for local explanations. For one, if we can gain a high degree of functional interpretability, our grasp of a predictor is global and should thus be preferred over local explanations of black-box predictions, other things being equal.\footnote{Other things encompassing, in particular, the accuracy of a predictor. It is stressed, e.g., by \citet{rudin2019} that the use of interpretable models will not depreciate accuracy in many cases and should therefore be preferred over xAI methods.} What is more, the discussion in the present paper suggests that obtaining visualizations of linear approximations is by no means the only way to gain interpretability. If the story presented here is correct, we can gain understanding of black-box models through a combination of understanding formal properties as well as visualizations. This is not to say that we will be able to use the methods from MARS, GAMs or trees on DNNs in a straightforward manner. Dealing with DNNs will require us to move beyond the linear paradigm and discussing how we can grasp complex interactions in high-dimensional settings.

\section{Conclusion}
\label{sec:conclusion}

Interpretability was analyzed by examining four cases of models with a certain degree of interpretability. There is a considerable heterogeneity with respect to the means by which we achieve \emph{functional interpretability}. In particular, linear models and decision trees belong to two different paradigms of how interpretability is achieved. The same is true for the two more general, but still interpretable models, MARS and GAMs. However, interpretability is not ill-defined for these reasons. Rather, we can spell out clearly what interpretability amounts to in particular cases.

The above discussion is limited in several respects and should be extended. First, the case of classification problems, and of different kinds of inputs (discrete, mixed etc.) should be taken into consideration.  Second, it would be desirable to extend the limited notion of functional interpretability explored here and related it to other existing analyses of interpretability, giving more weight to optimization, the representational role and the inner workings of models. Third, the investigation of interpretability should be extended to unsupervised learning, reinforcement learning, and other paradigms of ML.

\end{document}